\pdfoutput=1

\documentclass[11pt]{article}

\usepackage{emnlp2021}

\usepackage{times}
\usepackage{latexsym}
\usepackage{multirow}
\usepackage{graphicx}
\usepackage{algorithm}
\usepackage{algorithmic}
\usepackage{amsmath}
\usepackage{url}
\usepackage{comment}
\usepackage{enumitem}
\usepackage{color}
\usepackage{subfigure}
\usepackage[T1]{fontenc}
\usepackage{footnote}
\usepackage{hyperref}
\usepackage{listings}
\usepackage{booktabs}

\usepackage{xcolor}

\definecolor{codegreen}{rgb}{0,0.6,0}
\definecolor{codegray}{rgb}{0.5,0.5,0.5}
\definecolor{codepurple}{rgb}{0.58,0,0.82}
\definecolor{backcolour}{rgb}{0.95,0.95,0.92}

\lstdefinestyle{mystyle}{
    backgroundcolor=\color{backcolour},   
    commentstyle=\color{codegreen},
    stringstyle=\color{codepurple},
    basicstyle=\ttfamily\scriptsize,
    breakatwhitespace=true,         
    breaklines=true,                 
    captionpos=b,                    
    keepspaces=true,                 
    numbers=none,                    
    numbersep=5pt,                  
    showspaces=false,                
    showstringspaces=false,
    showtabs=false,                  
    tabsize=2,
    columns=flexible,
    escapeinside={(*}{*)},
}

\usepackage[utf8]{inputenc}

\usepackage{microtype}

\newcommand{\boldpara}[1]{\vspace{0.12cm}\noindent \textbf{#1}}

\title{EasyNLP: A Comprehensive and Easy-to-use Toolkit for \\ Natural Language Processing}

\author{Chengyu Wang$^1$, Minghui Qiu$^1$\thanks{ \quad Corresponding Author.}, Chen Shi$^1$, Taolin Zhang$^{1,2}$, Tingting Liu$^{1,2}$, Lei Li$^{1,2}$, \\
\textbf{Jianing Wang$^{1,2}$, Ming Wang$^1$, Jun Huang$^1$, Wei Lin$^1$}\\
  $^1$ Platform of AI (PAI), Alibaba Group
  $^2$ East China Normal University\\
  \texttt{\{chengyu.wcy,minghui.qmh,huangjun.hj\}@alibaba-inc.com}
}

\begin{document}
\maketitle
\begin{abstract}
The success of Pre-Trained Models (PTMs) has reshaped the development of Natural Language Processing (NLP). Yet, it is not easy to obtain high-performing models and deploy them online for industrial practitioners. To bridge this gap, EasyNLP is designed to make it easy to build NLP applications, which supports a comprehensive suite of NLP algorithms. 
It further features knowledge-enhanced pre-training, knowledge distillation and few-shot learning functionalities for large-scale PTMs, and provides a unified framework of model training, inference and deployment for real-world applications. 
Currently, EasyNLP has powered over ten business units within Alibaba Group and is seamlessly integrated to the Platform of AI (PAI) products on Alibaba Cloud. The source code of our EasyNLP toolkit is released at GitHub (\url{https://github.com/alibaba/EasyNLP}).
\end{abstract}

\section{Introduction}

The Pre-Trained Models (PTMs) such as BERT, GPT-3 and PaLM have achieved remarkable results in NLP. With the scale expansion of PTMs, the performance of NLP tasks has been continuously improved; thus, there is a growing trend of ultra-large-scale pre-training, pushing the scale of PTMs from millions, billions, to even trillions~\cite{DBLP:conf/naacl/DevlinCLT19,DBLP:conf/nips/BrownMRSKDNSSAA20,palm}.

However, the application of large PTMs in industrial scenarios is still a non-trivial problem. The reasons are threefold. i) Large PTMs are not always smarter and can make commonsense mistakes due to the lack of world knowledge~\cite{DBLP:conf/emnlp/PetroniRRLBWM19}.
Hence, it is highly necessary to make PTMs explicitly understand world facts by knowledge-enhanced pre-training.
ii) Although large-scale PTMs have achieved good results with few training samples, the problem of insufficient data and the huge size of models such as GPT-3 still restrict the usage of these models. Thus, few-shot fine-tuning BERT-style PTMs is more practical for online applications~\cite{DBLP:journals/corr/abs-2012-15723}.
iii) Last but not least, although large-scale PTMs have become an important part of the NLP learning pipeline,
the slow training and inference speed seriously affects online applications that require higher QPS (Query Per Second) with limited computational resources. 

To address these issues, we develop EasyNLP, an NLP toolkit that is designed to make the landing of large PTMs efficiently and effectively.
EasyNLP provides knowledge-enhanced pre-training functionalities to
improve the knowledge understanding abilities of PTMs. Specifically, our proposed DKPLM framework~\cite{DBLP:journals/corr/abs-2112-01047} enables 
the decomposition of knowledge-enhanced pre-training and task-specific learning. Hence, the resulting models can be tuned and deployed in the same way as BERT \cite{DBLP:conf/naacl/DevlinCLT19}.
EasyNLP also integrates a variety of popular prompt-based few-shot learning algorithms such as PET~\cite{DBLP:conf/eacl/SchickS21} and P-Tuning~\cite{DBLP:journals/corr/abs-2103-10385}.
Particularly, we propose a new few-shot learning paradigm named Contrastive Prompt Tuning (CP-Tuing)~\cite{DBLP:journals/corr/abs-2204-00166} that eases the manual labor of verbalizer construction based on contrastive learning.
Finally, EasyNLP supports several knowledge distillation algorithms that compress large PTMs into small and efficient ones.
Among them, the MetaKD algorithm~\cite{DBLP:conf/acl/Pan0QZLH20} can significantly improve the effectiveness of the learned models with  cross-domain datasets. 

Overall, our EasyNLP toolkit can provide users with large-scale and robust learning functionalities, and is seamlessly connected to the Platform of AI (PAI)\footnote{\url{https://www.alibabacloud.com/product/machine-learning}} products. Its rich APIs provide users with an efficient and complete experience from training to deployment of PTMs for various applications.

In a nutshell, the main features of the EasyNLP toolkit include the following aspects:
\begin{itemize}
\item\textbf{Easy-to-use and highly customizable.} In addition to providing easy-to-use commands to call cutting-edge NLP models, EasyNLP abstracts customized modules such as AppZoo and ModelZoo to make it easy to build NLP applications. It is seamlessly integrated to the PAI products on Alibaba Cloud, including PAI-DSW for model development, PAI-DLC for cloud-native training, PAI-EAS for online serving, and PAI-Designer for zero-code model training. It also features DataHub that provides users with a simple interface to load and process various types of NLP datasets.

\item\textbf{Knowledge-enhanced PTMs.}
EasyNLP is equipped with cutting-edge knowledge-enhanced PTMs of various domains.
Its pre-training APIs enables users to obtain customized PTMs using their own knowledge base with just a few lines of codes.

\item\textbf{Landing large-scale PTMs.} EasyNLP provides few-shot learning capabilities based on prompts, allowing users to fine-tune PTMs with only a few training samples to achieve good results. Meanwhile, it provides knowledge distillation functionalities to help quickly distill large models to small and efficient models to facilitate online deployment.

\item\textbf{Compatible with open-source community.} EasyNLP has rich APIs to support the training of models from other open-source libraries such as Huggingface/Transformers\footnote{\url{https://github.com/huggingface/transformers}} with the PAI's distributed learning framework. It is also compatible with the PTMs in EasyTransfer ModelZoo\footnote{\url{https://github.com/alibaba/EasyTransfer}}~\cite{DBLP:conf/cikm/QiuLWPWCJLH0021}.
\end{itemize}

\section{Related Work}

In this section, we summarize the related work on PTMs, prompt learning and knowledge distillation.


\subsection{Pre-trained Language Models}
PTMs have achieved significant improvements on various tasks by self-supervised pre-training~\cite{DBLP:journals/corr/abs-2003-08271}. 
To name a few, BERT~\cite{DBLP:conf/naacl/DevlinCLT19} learns bidirectional contextual representations by transformer encoders. 
Other transformer encoder-based PTMs include Transformer-XL~\cite{DBLP:conf/acl/DaiYYCLS19}, XLNet~\cite{DBLP:conf/nips/YangDYCSL19} and many others.
The encoder-decoder and auto-regressive decoder architectures are used in T5~\cite{DBLP:journals/jmlr/RaffelSRLNMZLL20} and GPT-3~\cite{DBLP:conf/nips/BrownMRSKDNSSAA20}.
Knowledge-enhanced PTMs~\citep{DBLP:conf/acl/ZhangHLJSL19,DBLP:conf/aaai/LiuZ0WJD020,DBLP:conf/coling/SunSQGHHZ20} improve language understanding abilities of PTMs via injecting relational triples extracted from knowledge bases.

\subsection{Prompt Learning for PTMs}
Prompt learning models the probability of texts directly as the model prediction results based on language models~\cite{DBLP:journals/corr/abs-2107-13586}. In the literature, PET~\cite{DBLP:conf/eacl/SchickS21} models NLP tasks as cloze problems and maps the results of the masked language tokens to class labels. \citet{DBLP:journals/corr/abs-2012-15723} generates discrete prompts from T5~\cite{DBLP:journals/jmlr/RaffelSRLNMZLL20} to support prompt discovery.
P-Tuning~\cite{DBLP:journals/corr/abs-2103-10385} learns continuous prompt embeddings with differentiable parameters.
Our CP-Tuing~\cite{DBLP:journals/corr/abs-2204-00166} optimizes the output results based on contrastive learning, without defining mappings from outputs to class labels.

\subsection{Knowledge Distillation}

Knowledge distillation aims at learning a smaller model from an ensemble or a larger model~\cite{DBLP:journals/corr/HintonVD15}. For large-scale PTMs, DistillBERT~\cite{DBLP:journals/corr/abs-1910-01108} and PKD~\cite{DBLP:conf/emnlp/SunCGL19} applies the distillation loss in the pre-training and fine-tuning stages, separately.
TinyBERT~\cite{DBLP:conf/emnlp/JiaoYSJCL0L20} further distills BERT in both stages, considering various types of signals. Due to space limitation, we do not further elaborate other approaches.
Our MetaKD method~\cite{DBLP:conf/acl/Pan0QZLH20} further improves the accuracy of the student models by exploiting cross-domain transferable knowledge, which is fully supported by EasyNLP.

\begin{figure*}[th!]
    \centering
	\includegraphics[width=\linewidth]{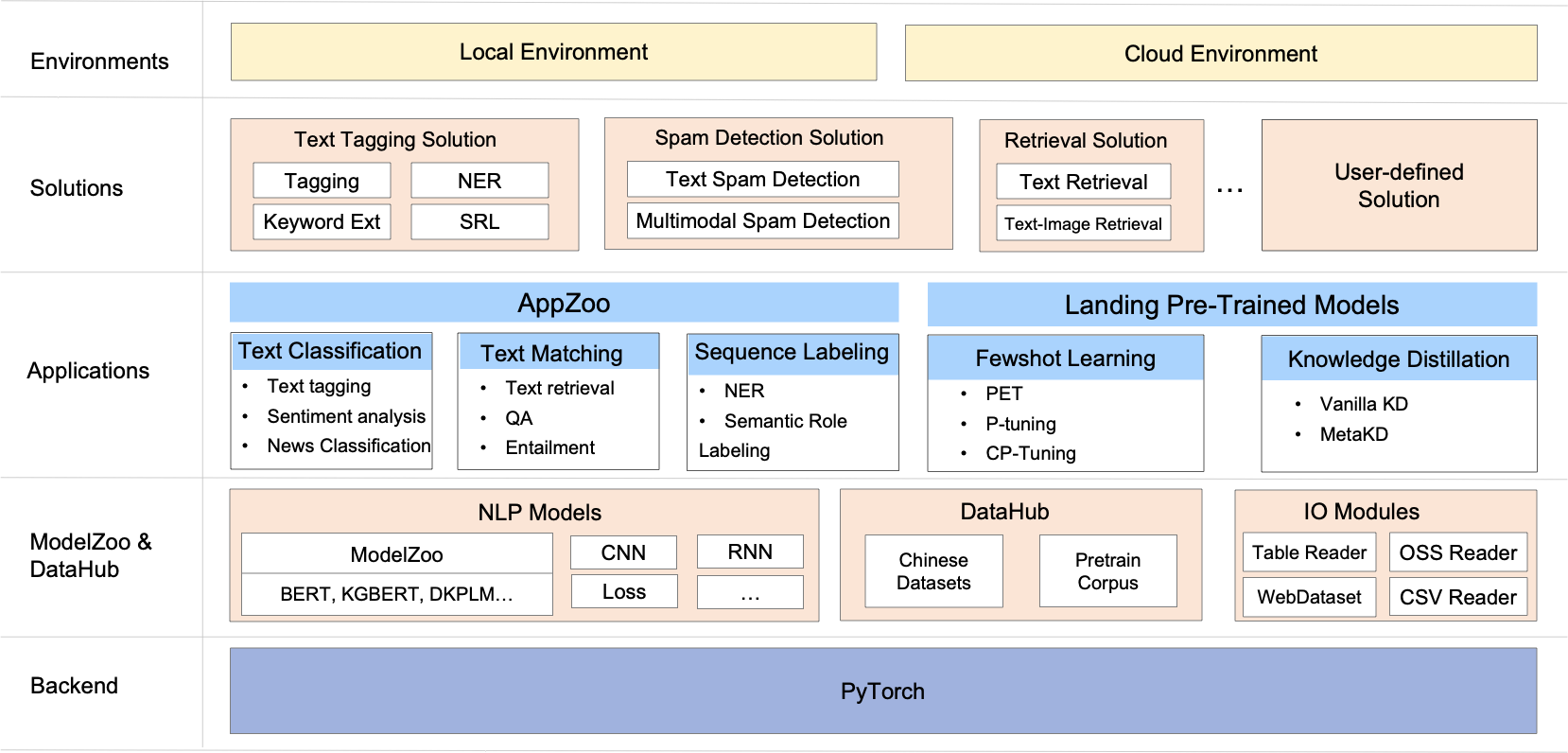}
	\caption{An overview of the EasyNLP toolkit.}\label{fig:overview}
\end{figure*}

\section{The EasyNLP Toolkit}

In this section, we introduce various aspects of our EasyNLP toolkit in detail.

\subsection{Overview}
We begin with an overview of EasyNLP in Figure~\ref{fig:overview}. EasyNLP is built upon PyTorch and supports rich data readers to process data from multiple sources. Users can load any PTMs from ModelZoo and datasets from DataHub, build their applications from AppZoo, or explore its advanced functionalities such as knowledge-enhanced pre-training, knowledge distillation and few-shot learning. The codes can run either in local environments or PAI's products on the cloud. In addition, all EasyNLP's APIs are also released to make it easy for users to customize any kinds of NLP applications.

\subsection{DataHub, ModelZoo and AppZoo}



\noindent\textbf{DataHub.} DataHub provides users with an interface to load and process various kinds of data. It is compatible with Huggingface datasets\footnote{\url{https://github.com/huggingface/datasets}} as a built-in library that supports unified interface calls and contains datasets of a variety of tasks.
Some examples are listed in Table~\ref{table:datahub}. 
Users can load the required data by specifying the dataset name through the \texttt{load\_dataset} interface, and then use the \texttt{GeneralDataset} interface to automatically process the data into model input. An example of loading and pre-processing the TNEWS dataset, together with its subsequent steps, is shown in Code~\ref{case1}. For user-defined datasets, it is also straightforward to inherit the \texttt{GeneralDataset} class to customize the data format.

\noindent\textbf{ModelZoo.} PTMs such as BERT \cite{DBLP:conf/naacl/DevlinCLT19}, RoBERTa \cite{DBLP:journals/corr/abs-1907-11692} and T5 \cite{DBLP:journals/jmlr/RaffelSRLNMZLL20} greatly improve the performance of NLP tasks. To facilitate user deployment of models, ModelZoo provides general pre-trained models as well as our own models for users to use, such as DKPLM~\cite{DBLP:journals/corr/abs-2112-01047} of various domains.
A few widely-used non-PTM models are also supported, such as Text-CNN~\cite{DBLP:conf/emnlp/Kim14}.


\newcounter{mpFootnoteValueSaver}
\setcounter{mpFootnoteValueSaver}{\value{footnote}}
\begin{table}
 \centering
 \begin{small}
 \begin{tabular}{l|l} 
  \toprule 
  \bf Task Type & \bf Example of Datasets \\ 
  \midrule 
     Sequence Classification & TNEWS\footnotemark, SogouCA\footnotemark \\
     Text Generation & THUCNews\footnotemark, SogouCS\footnotemark\\
     Few-shot / Zero-shot Learning & BUSTM\footnotemark, CHID\footnotemark \\
     Knowledge-based NLU & OntoNotes\footnotemark, SanWen\footnotemark \\
 \bottomrule 
 \end{tabular}
 \end{small}
 \caption{A partial list of datasets in EasyNLP DataHub.} 
\label{table:datahub}
\end{table}
\stepcounter{mpFootnoteValueSaver}%
    \footnotetext[\value{mpFootnoteValueSaver}]{\url{https://github.com/CLUEbenchmark/CLUE}}
\stepcounter{mpFootnoteValueSaver}%
    \footnotetext[\value{mpFootnoteValueSaver}]{\url{http://www.sogou.com/labs/resource/ca.php}}
\stepcounter{mpFootnoteValueSaver}%
    \footnotetext[\value{mpFootnoteValueSaver}]{\url{http://thuctc.thunlp.org/}}
\stepcounter{mpFootnoteValueSaver}%
    \footnotetext[\value{mpFootnoteValueSaver}]{\url{https://www.sogou.com/labs/resource/cs.php}}
\stepcounter{mpFootnoteValueSaver}%
    \footnotetext[\value{mpFootnoteValueSaver}]{\url{https://github.com/xiaobu-coai/BUSTM}}
\stepcounter{mpFootnoteValueSaver}%
    \footnotetext[\value{mpFootnoteValueSaver}]{\url{https://github.com/chujiezheng/ChID-Dataset}}
\stepcounter{mpFootnoteValueSaver}%
    \footnotetext[\value{mpFootnoteValueSaver}]{\url{https://catalog.ldc.upenn.edu/LDC2013T19}}
\stepcounter{mpFootnoteValueSaver}%
    \footnotetext[\value{mpFootnoteValueSaver}]{\url{https://github.com/lancopku/Chinese-Literature-NER-RE-Dataset}}

\begin{figure}
\begin{minipage}{0.5\textwidth}
\begin{lstlisting}[language=python, caption=Load the TNEWS training set and build a text classification application using EasyNLP., label=case1]
from easynlp.dataset import load_dataset, GeneralDataset

# load dataset
dataset = load_dataset('clue', 'tnews')["train"]
# parse data into classification model input
encoded = GeneralDataset(dataset, 'chinese-bert-base')
# load model
model = SequenceClassification('chinese-bert-base')
trainer = Trainer(model, encoded)
# start to train
trainer.train()
\end{lstlisting}
\end{minipage}
\end{figure}

\noindent\textbf{AppZoo.} To help users build NLP applications more easily with our framework, we further provide a comprehensive NLP application tool named AppZoo. It supports running applications with a few command-line arguments and provides a variety of mainstream or innovative NLP applications for users. AppZoo provides rich modules for users to build different application pipelines, including language modeling, feature vectorization, sequence classification, text matching, sequence labeling and many others.
An example of training a text classifier using AppZoo is shown in Code~\ref{case2}.

\begin{figure}
\begin{minipage}{0.5\textwidth}
\begin{lstlisting}[language=sh, caption=AppZoo for training a BERT-based text classifier using EasyNLP., label=case2]
easynlp \
   --mode=train \
   --worker_gpu=1 \
   --tables=train.tsv,dev.tsv \
   --input_schema=sent:str:1,label:str:1 \
   --first_sequence=sent \
   --label_name=label \
   --label_enumerate_values=0,1 \
   --checkpoint_dir=./classification_model \
   --epoch_num=1  \
   --sequence_length=128 \
   --app_name=text_classify \
   --user_defined_parameters=
    'pretrain_model_name_or_path=bert-small-uncased'
\end{lstlisting}
\end{minipage}
\end{figure}



\subsection{In-house Developed Algorithms}

In this section, we introduce in-house developed algorithms in EasyNLP.
All these algorithms have been tested in real-world applications.

\subsection{Knowledge-enhanced Pre-training}

Knowledge-enhanced pre-training improves the performance of PTMs by injecting the relational facts from knowledge bases. Yet, a lot of existing works require additional knowledge encoders during pre-training, fine-tuning and inference~\cite{DBLP:conf/acl/ZhangHLJSL19,DBLP:conf/aaai/LiuZ0WJD020,DBLP:conf/coling/SunSQGHHZ20}.

The proposed DKPLM paradigm~\cite{DBLP:journals/corr/abs-2112-01047} decomposes the knowledge injection process. For DKPLM, knowledge injection is only applied during pre-training, without introducing extra parameters as knowledge encoders, alleviating the significant computational burdens for users. Meanwhile, during fine-tuning and inference stages, our model can be utilized in the same way as that of BERT~\cite{DBLP:conf/naacl/DevlinCLT19} and other plain PTMs, which facilitates the model fine-tuning and deployment in EasyNLP and other environments.
Specifically, the DKPLM framework introduces three novel techniques for knowledge-enhanced pre-training. It recognizes long-tail entities from text corpora for knowledge injection only, avoiding learning too much redundant and irrelevant information from knowledge bases~\cite{DBLP:conf/ijcai/ZhangDCCZZC21}. 
Next, the representations of entities are replaced by ``pseudo token representations'' derived from knowledge bases, without introducing any extra parameters to DKPLM. Finally, a relational knowledge decoding task is introduced to force the model to understand what knowledge is injected.

In EasyNLP, we provide the entire pre-training pipeline of DKPLM for users. In addition, a collection of pre-trained DKPLMs for specific domains have been registered in ModelZoo for supporting domain-specific applications.

\subsection{Few-shot Learning for PTMs}

For low-resource scenarios, prompt-based learning leverages prompts as task guidance for effective few-shot fine-tuning. In EasyNLP, to facilitate easy few-shot learning, we integrate PET~\cite{DBLP:conf/eacl/SchickS21} and P-Tuning~\cite{DBLP:journals/corr/abs-2103-10385} into AppZoo that allow users call the algorithms in the similar way compared to standard fine-tuning.

It should be further noted that either PET or P-Tuning require the explicit handcraft of verbalizers, which is a tedious process and may lead to unstable results. Our CP-Tuning approach~\cite{DBLP:journals/corr/abs-2204-00166} enables few-shot fine-tuning PTMs without the manual engineering of task-specific prompts and verbalizers. 
A pair-wise cost-sensitive contrastive learning is introduced to achieve verbalizer-free class mapping by learning to distinguish different classes.
Users can also explore CP-Tuning in AppZoo for any tasks that classical prompt-based methods support.

\subsection{Knowledge Distillation for PTMs}

The large model size and the long inference time hinder the deployment of large-scale PTMs to resource-constrained applications. In EasyNLP, we provide a complete learning pipeline for knowledge distillation, including data augmentation for training sets, logits extraction from teacher models and distilled training of student models.

In addition, we notice that a majority of existing approaches focus on a single domain only. The proposed MetaKD algorithm~\cite{DBLP:conf/acl/Pan0QZLH20} explicitly leverages the cross-domain transferable knowledge to improve the accuracy of student models. It first obtain a meta-teacher model to capture transferable knowledge at both instance-level and feature-level
from multiple domains.
Next, a meta-distillation algorithm is employed to learn single-domain student models with selective signals from the meta-teacher. In EasyNLP, the MetaKD process is implemented as a general feature for any types of BERT-style PTMs.

\begin{table*}
 \centering
 \begin{small}
 \begin{tabular}{l cccccccc} 
  \toprule 
 \bf PTM   & AFQMC & CMNLI & CSL & IFLYTEK & OCNLI & TNEWS & WSC & \bf Average \\ 
  \midrule 
BERT-base & 72.17 & 75.74 & 80.93 & 60.22 & 78.31 & 57.52 & 75.33 & 71.46 \\
BERT-large & 72.89 & 77.62 & 81.14 & 60.70 & 78.95 & 57.77 & 78.18 & 72.46 \\
RoBERTa-base & 73.10 & 80.75 & 80.07 & 60.98 & 80.75 & 57.93 & 86.84 & 74.35 \\
RoBERTa-large & 74.81 & 80.52 & 82.60 & 61.37 & 82.49 & 58.54 & 87.50 & 75.40 \\
MacBERT-base & 74.23 & 80.65 & 81.70 & 61.14 & 80.65 & 57.65 & 80.26 & 73.75 \\
MacBERT-large & 74.37 & 81.19 & 83.70 & 62.05 & 81.65 & 58.45 & 86.84 & 75.46 \\
 \bottomrule 
 \end{tabular}
 \end{small}
\caption{CLUE performance of BERT, RoBERTa and MacBERT fine-tuned with EasyNLP (\%).}
\label{table:benchmark-clue}
\end{table*}

\begin{table*}
\centering
\begin{small}
\begin{tabular}{l ccccccccc}
\toprule
\bf PTM             & MNLI & QNLI & QQP  & RTE  & SST-2 & MRPC & CoLA & STSB & \bf Average \\
\midrule
BERT-base         & 84.8 & 91.4 & 91.1 & 68.3 & 92.5  & 88.1 & 55.3 & 89.6 & 82.6                 \\
BERT-large        & 86.6 & 92.4 & 91.2 & 70.8 & 93.4  & 88.2 & 61.1 & 90.1 & 84.2                 \\
RoBERTa-base      & 87.3 & 92.5 & 92.1 & 77.3 & 94.9  & 90.2 & 63.9 & 91.1 & 86.2                 \\
RoBERTa-large     & 90.1 & 94.5 & 92.3 & 87.1 & 96.4  & 91.0 & 67.8 & 92.3 & 88.9                 \\ \bottomrule 
\end{tabular}
\end{small}
\caption{GLUE performance of BERT and RoBERTa fine-tuned with EasyNLP (\%).}
\label{table:glue}
\end{table*}

\begin{table*}
\centering
\begin{small}
\begin{tabular}{l|ccc|ccc|c}
\toprule
Dataset & ELMo & BERT & RoBERTa &  CoLAKE & K-Adapter$^\ast$ & KEPLER & DKPLM \\
\midrule
Google-RE & 2.2\% & 11.4\% & 5.3\%  & 9.5\%  & 7.0\% & 7.3\% & \textbf{10.8}\% \\
UHN-Google-RE & 2.3\% & 5.7\% & 2.2\% & 4.9\%  & 3.7\% & 4.1\% & \textbf{5.4}\% \\ \midrule
T-REx & 0.2\%  & 32.5\% & 24.7\% & 28.8\%  & 29.1\% & 24.6\% & \textbf{32.0}\% \\
UHN-T-REx & 0.2\% & 23.3\% & 17.0\% & 20.4\% & \textbf{23.0}\%  & 17.1\% & 22.9\% \\ \bottomrule
\end{tabular}
\end{small}
\caption{The performance on LAMA knowledge probing datasets.
Note that K-Adapter is trained based on a large-scale model and uses a subset of T-REx as its training data.}
\label{knowledge_probing_result}
\end{table*}

\section{System Evaluations and Applications}
\label{sec:exp}
In this section, we empirically examine the effectiveness and efficiency of the EasyNLP toolkit on both public datasets and industrial applications.

\subsection{CLUE and GLUE Benchmarks}

In order to validate the effectiveness of EasyNLP on model fine-tuning, we fine-tune PTMs on the CLUE and GLUE benchmarks~\cite{DBLP:conf/iclr/WangSMHLB19,DBLP:conf/coling/XuHZLCLXSYYTDLS20}. For all tasks, we use a limited hyper-parameter
search space, with batch sizes in $\{8, 16, 32, 48\}$, sequence length in $\{128,256\}$ and learning rates in $\{1e-5, 2e-5, 3e-5, 4e-5, 5e-5\}$.
The underlying PTMs include BERT~\cite{DBLP:conf/naacl/DevlinCLT19} and RoBERTa~\cite{DBLP:journals/corr/abs-1907-11692}. We also evaluate MacBERT~\cite{DBLP:conf/emnlp/CuiC000H20} for the Chinese benchmark CLUE.
We report the results over the development sets of each task in the two benchmarks, shown in Tables~\ref{table:benchmark-clue} and~\ref{table:glue}, respectively.
The obtained comparable performance has shown the reliability of EasyNLP, which achieves similar performance compared to other open-source frameworks and their original implementations.

\begin{table*}
\centering
\begin{small}
\begin{tabular}{l | ccccccccc}
\toprule 
\textbf{Method} & SST-2 & MR & CR & MRPC & QQP & QNLI & RTE & SUBJ & \bf Avg.\\
 \midrule 
 Standard Fine-tuning & 78.62	& 76.17 & 72.48 & 64.40 & 63.01 & 62.32	& 52.28 & 86.82 & 69.51\\
 PET & 92.06 & 87.13 & 87.13 & 66.23 & 70.34 & 64.38 & 65.56 & 91.28 & 78.01\\
 LM-BFF (Auto T) & 90.60 & 87.57 & 90.76 & 66.72 & 65.25 & 68.87 & 65.99 & 91.61 & 78.42\\
 LM-BFF (Auto L) & 90.55 & 85.51 & 91.11 & 67.75 & 70.92 & 66.22 & 66.35 & 90.48 & 78.61\\
 LM-BFF (Auto T+L) & 91.42 & 86.84 & 90.40 & 66.81 & 61.61 & 61.89 & 66.79 & 90.72 & 77.06\\
 P-tuning & 91.42 & 87.41 & 90.90 & 71.23 & 66.77 & 63.42 & 67.15 & 89.10 & 78.43\\
 \midrule 
 \bf CP-Tuning & \bf 93.35 & \bf 89.43 & \bf 91.57 & \bf 72.60 & \bf 73.56 &  \bf 69.22 & \bf 67.22 & \bf 92.27 & \bf 81.24\\
 \bottomrule 
\end{tabular}
\end{small}
\caption{Comparison between CP-Tuning and baselines over the testing sets in terms of accuracy (\%).}
\label{tab:few-shot}
\end{table*}

\begin{table}
 \centering
 \begin{small}
 \begin{tabular}{l cc} 
  \toprule 
\bf Method  & Amazon & MNLI\\ 
  \midrule 
$\text{BERT}\text{-s}$ & 87.9 & 81.9 \\
$\text{BERT}\text{-mix}$ & 89.5 & 84.4 \\
$\text{BERT}\text{-mtl}$ & 89.8 & 84.2 \\
 \midrule
$\text{BERT}\text{-s}$ $\rightarrow$ TinyBERT & 86.7 & 79.3\\
$\text{BERT}\text{-mix}$ $\rightarrow$ TinyBERT & 87.3 & 79.6\\
$\text{BERT}\text{-mtl}$ $\rightarrow$ TinyBERT &  87.7 & 79.7\\
 \midrule
\bf MetaKD & \textbf{89.4} & \textbf{80.4}\\
 \bottomrule 
  \end{tabular}
  \end{small}
  \caption{Evaluation of MetaKD over Amazon reviews and MNLI in terms of averaged accuracy (\%).} 
\label{table:results-amazon}
\end{table}

\subsection{Evaluation of Knowledge-enhanced Pre-training}
We report the performance of DKPLM over zero-shot knowledge probing tasks, including LAMA \cite{DBLP:conf/emnlp/PetroniRRLBWM19} and LAMA-UHN \cite{DBLP:journals/corr/abs-1911-03681}, with the results summarized in Table~\ref{knowledge_probing_result}.
Compared to strong baselines (i.e., CoLAKE \cite{DBLP:conf/coling/SunSQGHHZ20} K-Adapter \cite{DBLP:conf/acl/WangTDWHJCJZ21} and KEPLER \cite{DBLP:journals/corr/abs-1911-06136}), we see that
DKPLM achieves state-of-the-art results over three datasets (+1.57\% on average). The result of DKPLM is only 0.1\% lower than K-Adapter, without using any T-REx training data and larger backbones.
We can see that our pre-training process based on DKPLM can effectively store and understand factual relations from knowledge bases.

\boldpara{Industrial Applications.}
Based on the proposed DKPLM framework~\cite{DBLP:journals/corr/abs-2112-01047}, we have pre-trained a series of domain-specific PTMs to provide model service inside Alibaba Group, such as medical and finance domains, and observed consistent improvement in downstream NLP tasks. For example, the medical-domain DKPLM improves the accuracy of a medical named entity recognition task by over 3\%, compared to the standard BERT model~\cite{DBLP:conf/naacl/DevlinCLT19}. The pre-trained model (named~\texttt{pai-dkplm-medical-base-zh}) has also been released in our EasyNLP ModelZoo.

\subsection{Evaluations of Few-shot Learning}

We compare CP-Tuning~\cite{DBLP:journals/corr/abs-2204-00166} against several prompt-based fine-tuning approaches including PET~\cite{DBLP:conf/eacl/SchickS21}, LM-BFF~\cite{DBLP:journals/corr/abs-2012-15723} (in three settings where ``Auto T'', ``Auto L'' and ``Auto T+L'' refer
to the prompt-tuned PTM with automatically generated templates, label
words and both, respectively) and P-Tuning~\cite{DBLP:journals/corr/abs-2103-10385}. The experiments are conducted over several text classification datasets in a 16-shot learning setting. The underlying PTM is RoBERTa~\cite{DBLP:journals/corr/abs-1907-11692}. Readers can refer to~\citet{DBLP:journals/corr/abs-2204-00166} for more details.
From the results in Table~\ref{tab:few-shot}, we can see that 
the performance gains of CP-Tuning over all the tasks
are consistent, compared to state-of-the-art methods.

\boldpara{Industrial Applications.}
For business customer service, it is necessary to extract the fine-grained attributes and entities from texts, which may involve a large number of classess with few training data available. By applying our algorithm in EasyNLP, the accuracy scores of entity and attribute extraction are improved by 2\% and 5\%. In addition, our few-shot toolkit produces the best performance on the FewCLUE benchmark~\cite{DBLP:conf/nlpcc/XuWLLWHQTH21}.

\subsection{Evaluations of Knowledge Distillation}

We further report the performance of MetaKD~\cite{DBLP:conf/acl/Pan0QZLH20} on Amazon reviews~\citep{DBLP:conf/acl/BlitzerDP07} and MNLI~\cite{DBLP:conf/naacl/WilliamsNB18}, where the two datasets contain four and five domain instances, respectively.
In the experiments, we train the meta-teacher over multi-domain training sets, and distill the meta-teacher to each of all the domains.
The teacher model is BERT-base (with 110M parameters), while the student model is BERT-tiny (with 14.5M parameters).
Table~\ref{table:results-amazon} shows the performance of baselines and MetaKD, in terms of averaged accuracy across domains. BERT-s refers to a single BERT teacher trained on each domain. BERT-mix is one BERT teacher trained on the mixture of all domain data. BERT-mtl is one teacher trained by multi-task learning over all domains. For distillation, ``$\rightarrow$ TinyBERT'' means using the KD method described in~\citet{DBLP:journals/corr/abs-1909-10351} to distill the corresponding teacher model.
The results show that MetaKD significantly reduces the model size while preserving a similar performance. For more details, we refer the readers to~\citet{DBLP:conf/acl/Pan0QZLH20}.

\boldpara{Industrial Applications.} Distilled PTMs have been widely used inside Alibaba Group due to the high QPS requirements of online e-commerce applications. For example, in the AliMe chatbot~\cite{DBLP:conf/acl/QiuLWGCZCHC17}, we distill the BERT-based query intent detection model from the base version to the tiny version, resulting in 7.2x inference speedup while the accuracy is only decreased by 1\%.

\section{Conclusion}

In this paper, we introduced EasyNLP, a toolkit that is designed to make it easy to develop and deploy deep NLP applications based on PTMs. 
It supports a comprehensive suite of NLP algorithms and features knowledge-enhanced pre-training, knowledge distillation and few-shot learning functionalities for large-scale PTMs. 
Currently, EasyNLP has powered a number of business units inside Alibaba Cloud and provided NLP service on the cloud.
The toolkit has been open-sourced to promote research and development for NLP applications.

\newpage

\section*{Acknowledgments}
We thank Haojie Pan, Peng Li, Boyu Hou, Xiaoqing Chen, Xiaodan Wang, Xiangru Zhu and many other members of the Alibaba PAI team for their contribution and suggestions on building the EasyNLP toolkit.


\end{document}